\documentclass{article} 

\usepackage{iclr2017_conference,times}
\usepackage{hyperref}
\usepackage{url}

\usepackage{array}
\usepackage{mathtools}
\usepackage{amsmath, amssymb, amsthm}
\usepackage{color}
\usepackage{graphicx} 
\usepackage{subfigure}
\usepackage{algorithm}
\usepackage{algorithmic}
\usepackage{enumitem}
\usepackage{booktabs}
\usepackage{changepage}
\usepackage{bbm}
\usepackage{kky}
\usepackage{bartoDefns}

\newcommand{\modifone}[1]{#1}

\title{Batch Policy Gradient  Methods for \\  Improving Neural Conversation Models}

\newcommand{\affilsignkk}{$\,^\ast$}
\newcommand{\affilsignyb}{$\,^\flat$}

\author{Kirthevasan Kandasamy\affilsignkk \\
Carnegie Mellon University, 
Pittsburgh, PA, USA \\
\texttt{kandasamy@cs.cmu.edu} 
\And
Yoram Bachrach\affilsignyb \\
DigitalGenius Ltd., London, UK \\
\texttt{yorambac@gmail.com} 
\And
Ryota Tomioka, $\,$ Daniel Tarlow, $\,$ David Carter \\
Microsoft Research, Cambridge, UK \\
\texttt{\{ryoto,dtarlow,dacart\}@microsoft.com} \\[0.05in]
$\,$\affilsignkk\affilsignyb$\,$
{\footnotesize
This work was done when KK/YB was an intern/employee at Microsoft Research, Cambridge, UK.
\vspace{-0.15in}
}
}

\iclrfinalcopy 

\begin{document}
\pdfoutput=1


\newcommand{\imarrwthree}{1.80in}
\newcommand{\imhspthree}{-0.08in}
\newcommand{\imtextspace}{-0.10in}
\newcommand{\imcaptionspace}{-0.25in}

\newcommand{\insertFigSynOne}{
\begin{figure*}
\centering
\hspace{\imhspthree}
\subfigure[]{
  \includegraphics[width=\imarrwthree]{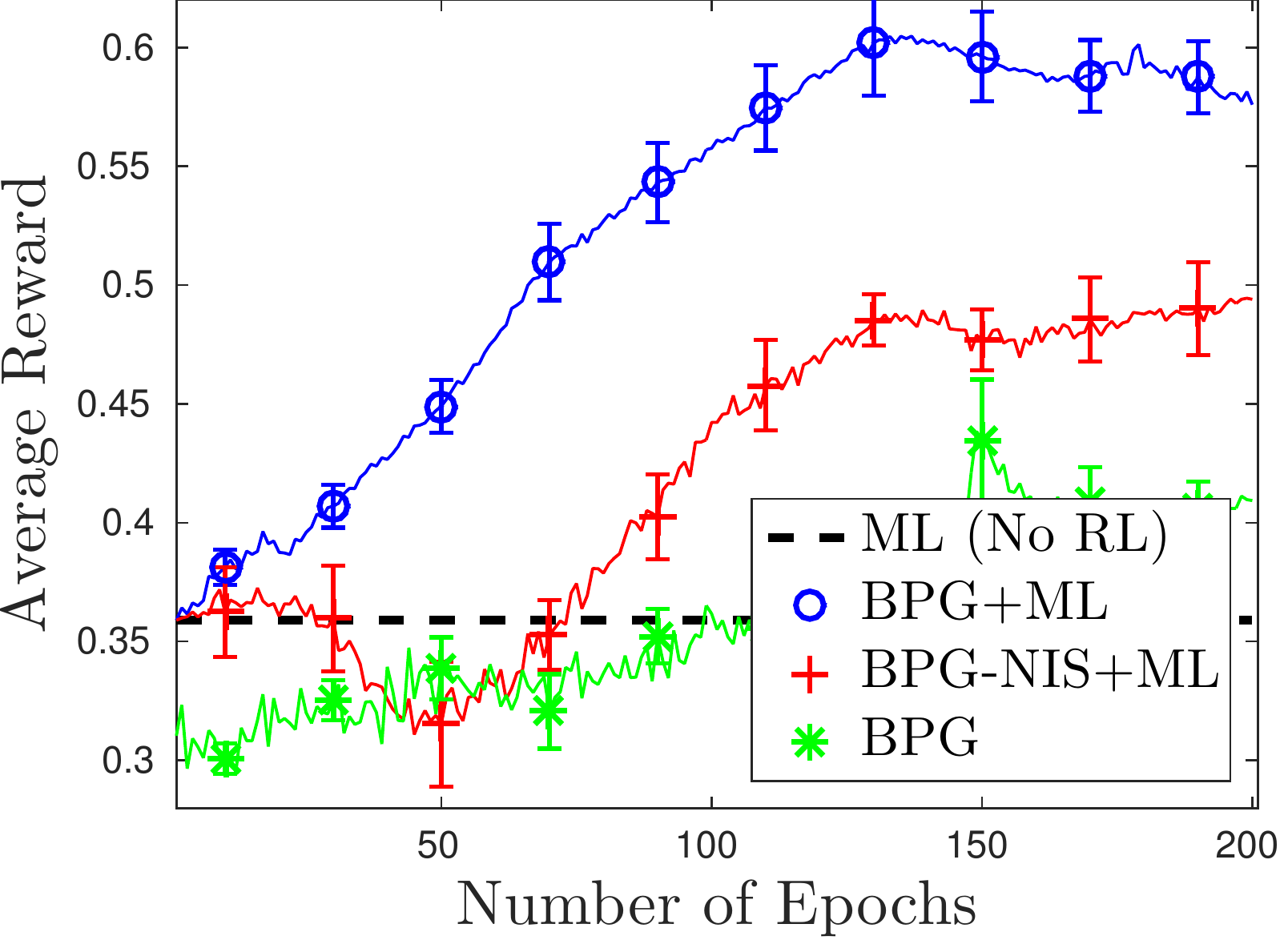} \hspace{\imhspthree}
  \label{fig:ml}
}
\subfigure[]{
  \includegraphics[width=\imarrwthree]{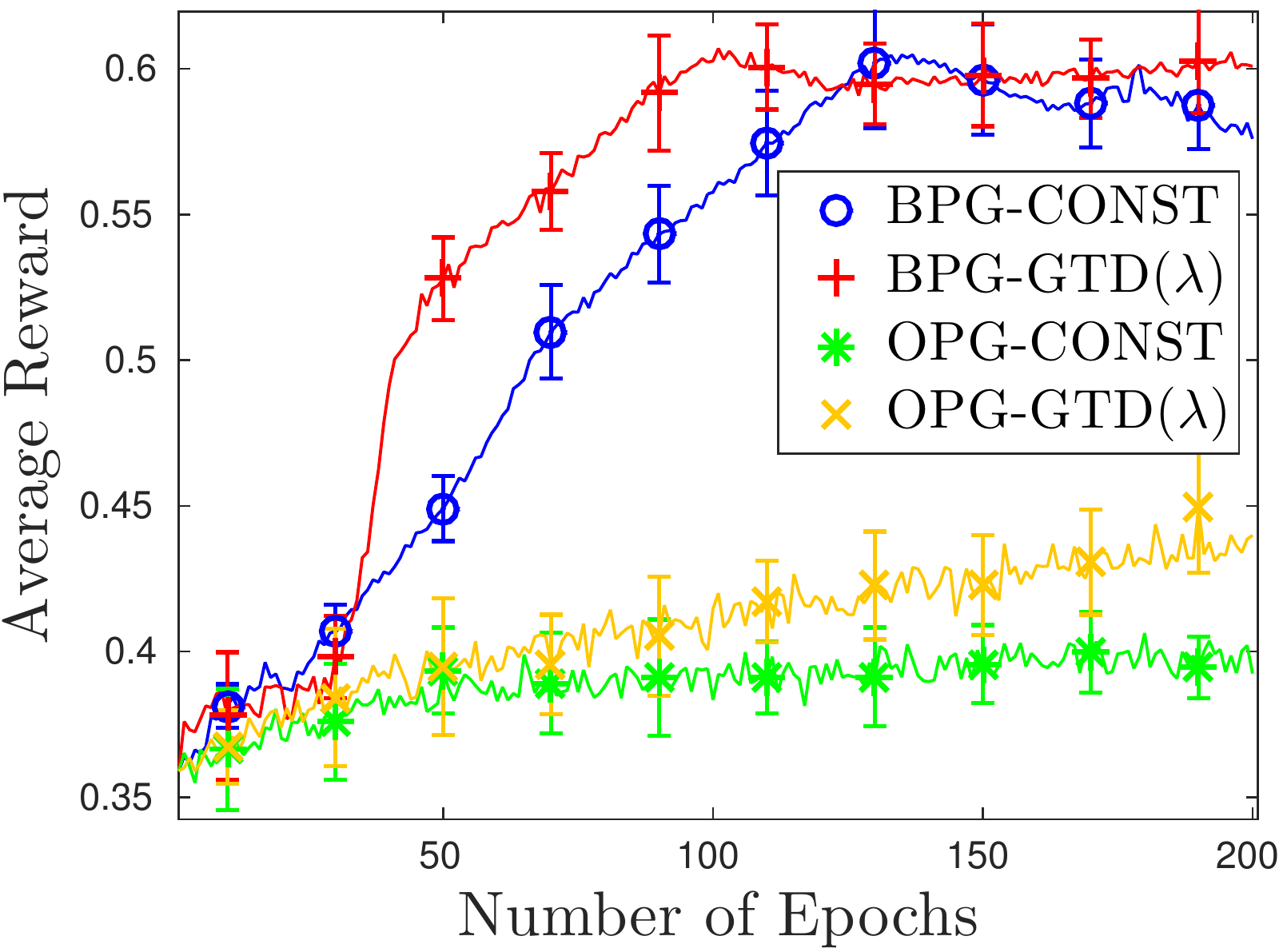} \hspace{\imhspthree}
  \label{fig:constvsgtdl}
}
\subfigure[]{
  \includegraphics[width=\imarrwthree]{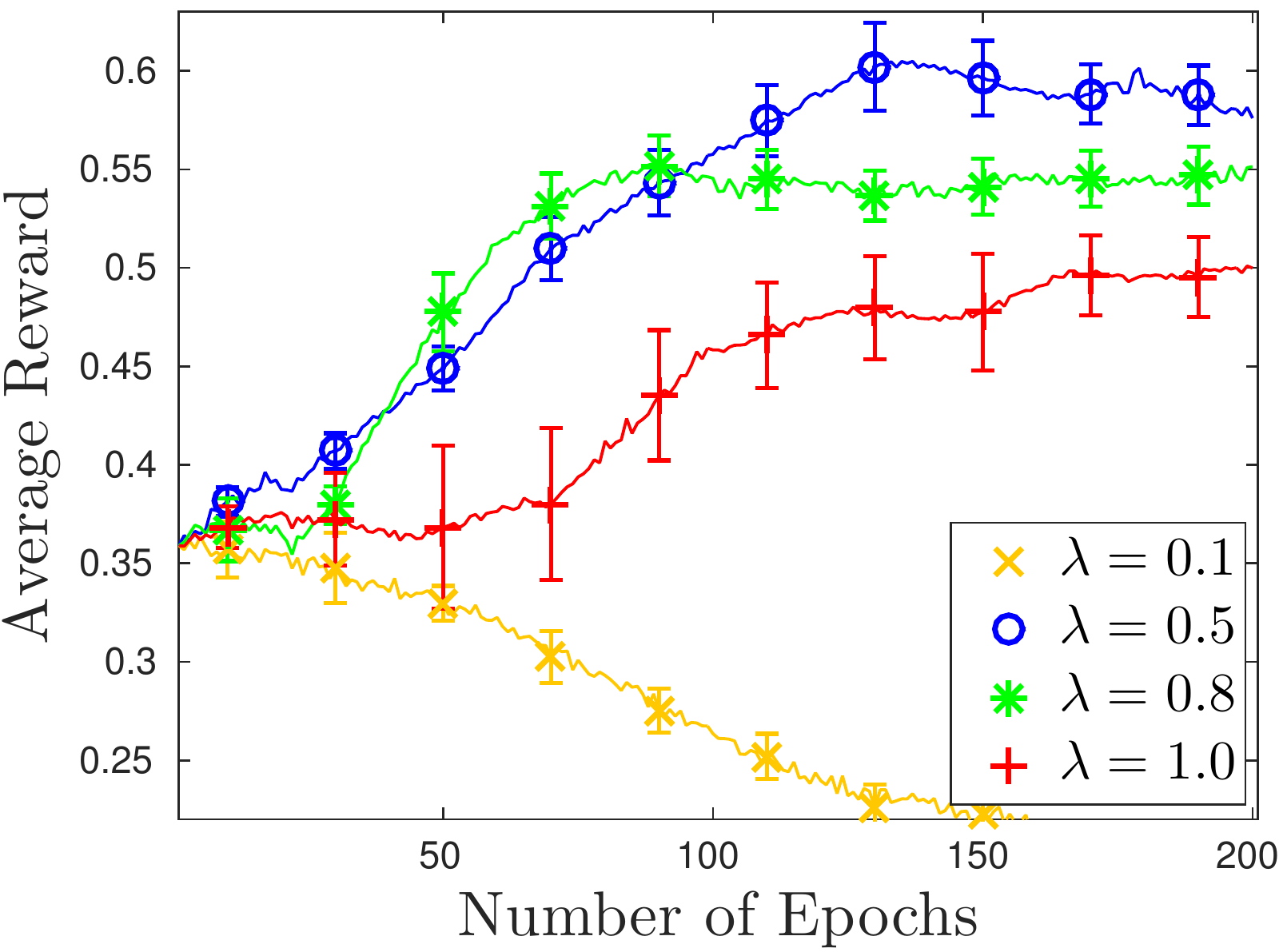} \hspace{\imhspthree}
  \label{fig:difflambda}
}
\vspace{\imcaptionspace}
\caption[]{\small
\label{fig:synresults}
Results for synthetic experiments.
~\subref{fig:ml}: Comparison of \bpgs with and without
maximum likelihood (\ml) initialisation and \bpgs without importance
sampling (\bpgnis).
The dotted line indicates performance of \mls alone.
~\subref{fig:constvsgtdl}: Comparison of \bpgs with its online counterparts \opg.
We compare both methods using a constant estimator (\const) for the value function
and \gtdl.
~\subref{fig:difflambda}: Comparison of \bpgs with different values of $\lambda$.
All curves were averaged over $10$ experiments where the training set was picked randomly
from a pool. The test set was the same in all $10$ experiments.
The error bars indicate one standard error.
\vspace{\imtextspace}
}
\end{figure*}
}

\newcommand{\insertFigLSTM}{
\begin{figure}
\centering
  \vspace{-0.2in}
  \begin{minipage}[c]{3.3in}
    \includegraphics[width=3.4in]{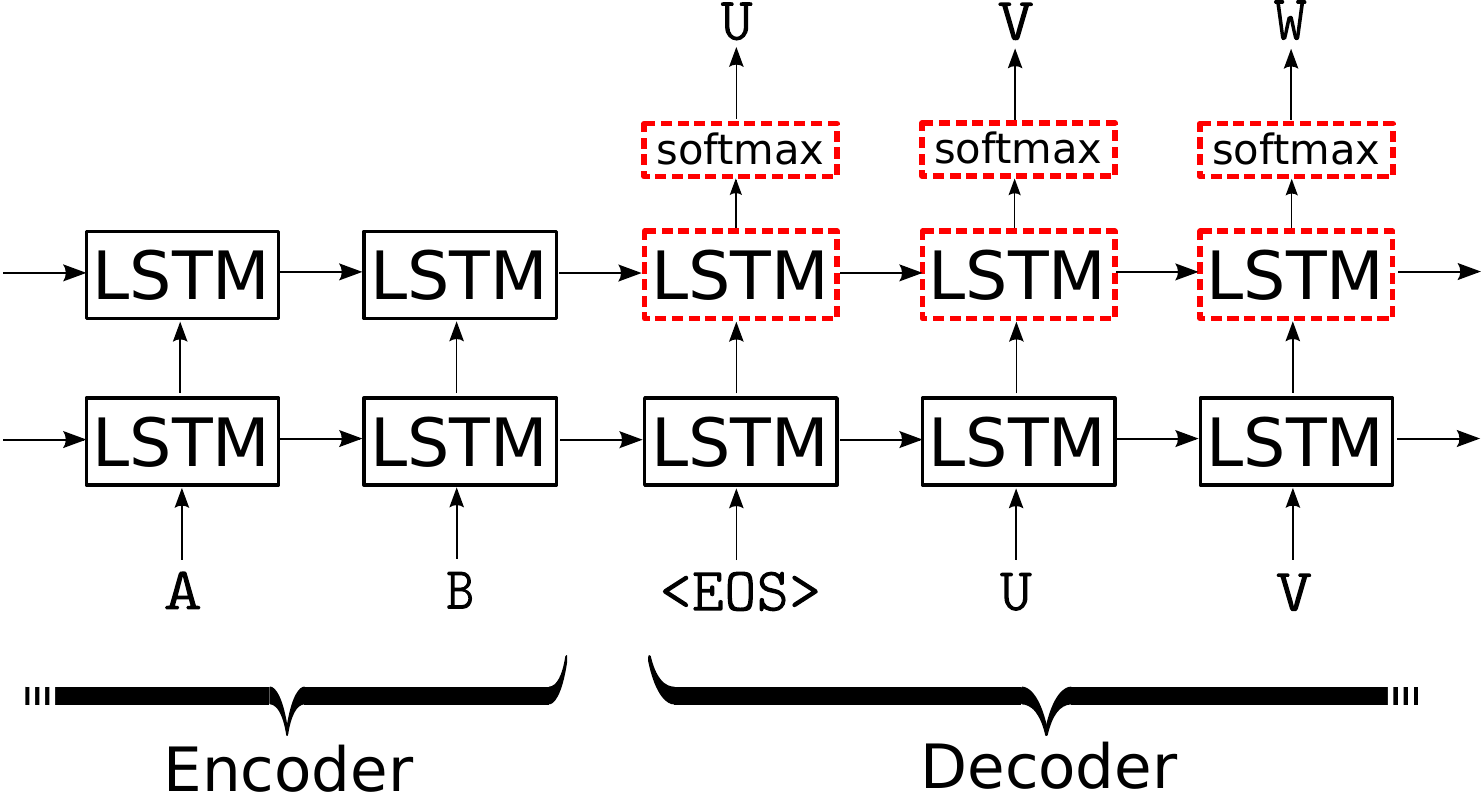}
  \end{minipage} \hspace{0.2in}
  \begin{minipage}[l]{1.9in}
  \vspace{-0.3in}
    \caption{\small
    Illustration of the encoder and decoder RNNs used in our experiments.
    In this example, the input to the encoder  is
    $x=(...,\tokenchars{A}, \tokenchars{B},\eostoken)$ and the output of the decoder is
    $y=(\tokenchars{U}, \tokenchars{V}, \tokenchars{W}, \dots)$.
    We use four different LSTMs for the bottom and top layers of the encoder and
    decoder networks.
    In our RL algorithms, we only change the top LSTM and the
    softmax layer of the decoder RNN as shown in red dashed lines.
    } 
\label{fig:lstmIllus}
  \end{minipage}
  \vspace{\imtextspace}
\end{figure}
}

\newcommand{\insertTableLPG}{
\newcommand{\bulletone}{--}
\newcommand{\bullettwo}{{\small $\blacktriangleright$}}
\newcommand{\bulletthree}{$\bullet$}
\newcommand{\bulletfour}{-}
\begin{algorithm}
\textbf{Given:} $\;$ Data $\{(x_i, y_i, r_i)\}_{i=1}^n$,
  step size $\alpha$, 
  return coefficient $\lambda$,
  initial $\theta_0$.
\begin{itemize}[leftmargin=0.3in]
\item[\bulletone] Set $\;\;\theta\leftarrow\theta_0$.
\item[\bulletone] For each epoch $k = 1, 2, \dots$
  \begin{itemize}
  \item[\bullettwo] Set $\;\;\Delta \theta \leftarrow \zero$
  \item[\bullettwo] For each episode $i = 1,\dots,n$
    \begin{itemize}
    \item[\bulletthree] $\rlambdaTpo \leftarrow r_i$
    \item[\bulletthree] $\rhot \leftarrow \policytheta(\ait|\sit)/q(\ait|\sit)$ for 
      $t=1,\dots,\Ti$.
    \item[\bulletthree] For each time step in \emph{reverse} $t= \Ti,\dots, 1$
      \begin{itemize}[leftmargin=0.4in]
      \item[(i)] 
        $\rlambdat \leftarrow (1-\lambda)\valfunhat(\sitpo) + \lambda \rhot \rlambdatpo$
      \item[(ii)] 
          $\Delta\theta \leftarrow  \Delta\theta \,+\,
          \frac{1}{\Ti}\rhot\psi(\sit,\ait)(\rlambdat - \valfunhat(\sit))$
      \item[(iii)] 
          Compute updates for the value function estimate $\valfunhat$.
      \end{itemize}
    \end{itemize}
  \item[\bullettwo] Update the policy
          $\quad\theta \leftarrow \theta + \alpha \Delta\theta$
  \item[\bullettwo] Update the value function estimate $\valfunhat$.
  \end{itemize}
\end{itemize}
\caption{$\quad$ Batch Policy Gradient~(\bpg)
\label{alg:logpolgrad}
}
\end{algorithm}
}

\newcommand{\insertRealResultsFeb}{
\begin{table}
\begin{center}
\begin{tabular}{l|c|c|c|c}
\toprule
$\;$ & Mean (\ml) & Mean (\bpgml) & Paired t-test & Wilcoxon  \\
\midrule
{Bot-1} & $0.8951 \pm 0.0070$ & $0.9052 \pm 0.0069$ & $0.10296$  & 0.07930 \\
{Bot-2} & $0.7009 \pm 0.0066$ & $0.7317 \pm 0.0066$ & $0.00007$  & 0.00017 \\
\bottomrule
\end{tabular}
\vspace{-0.1in}
\end{center}
\caption{\small
\label{tb:mturk}
\small
The results on the Mechanical Turk experiments using the restaurant dataset.
The first two columns are the mean labels of all responses before and after applying 
\bpgs on the bots initialised via maximum likelihood.
The last two columns are the p-values using a paired t-test and a paired
Wilcoxon signed rank test.
For both Bot-1 and Bot-2, we obtained 16,808  before and after responses scored by the
same worker.
Bot-2 is statistically significant at the $10\%$ level on both tests while Bot-1
is significant on the Wilcoxon test.
}
\vspace{-0.1in}
\end{table}
}

\newcommand{\insertRealResultsNew}{
\begin{table}
\begin{center}
\begin{tabular}{l|c|c|c|c|c}
\toprule
$\;$ & Mean (\ml) & Mean (\bpgml) & Paired t-test & Wilcoxon & Better / Worse  \\
\midrule
{Bot-1} & $0.8951 \pm 0.0070$ & $0.9052 \pm 0.0069$ & $0.10296$  & 0.07930 & 3916 / 2620 \\
{Bot-2} & $0.7009 \pm 0.0066$ & $0.7317 \pm 0.0066$ & $0.00007$  & 0.00017 & 4167 / 2867 \\
\bottomrule
\end{tabular}
\vspace{-0.1in}
\end{center}
\caption{\small
\label{tb:mturk}
\small
The results on the Mechanical Turk experiments using the restaurant dataset.
The first two columns are the mean labels of all responses before and after applying 
\bpgs on the bots initialised via maximum likelihood.
The last two columns are the p-values using a paired t-test and a paired
Wilcoxon signed rank test.
The last column shows the number of responses in with the \bpg--bot had better and worse
scores than the \ml--bot.
For both Bot-1 and Bot-2, we obtained 16,808  before and after responses scored by the
same worker.
Bot-2 is statistically significant at the $10\%$ level on both tests while Bot-1
is significant on the Wilcoxon test.
}
\vspace{-0.1in}
\end{table}
}

\newcommand{\insertRealResults}{
\begin{table}
\begin{center}
\begin{tabular}{l|c|c|c|c}
\toprule
$\;$ & Mean (\ml) & Mean (\bpgml) & Paired t-test & Wilcoxon signed rank  \\
\midrule
{Bot-1} & $0.8941 \pm 0.0103$ & $0.9066 \pm 0.0104$ & $0.15022$  & 0.23463 \\
{Bot-2} & $0.7052 \pm 0.0098$ & $0.7216 \pm 0.0099$ & $0.08735$  & 0.06286 \\
\bottomrule
\end{tabular}
\vspace{-0.1in}
\end{center}
\caption{\small
\label{tb:mturk}
\small
The results on the Mechanical Turk experiments using the restaurant dataset.
The first two columns are the mean labels of all responses before and after applying 
\bpgs on the bots initialised via maximum likelihood.
The last two columns are the p-values using a paired t-test and a paired
Wilcoxon signed rank test.
For both Bot-1 and Bot-2 the before and after responses were scored by the same worker.
The results above are on $7439$ such scores for each bot.
Bot-2 is statistically significant at the $10\%$ level on both tests.
}
\vspace{-0.1in}
\end{table}
}

\newcommand{\postertablefontsz}{0.1pt}
\newcommand{\postertablebaselineskip}{0.2pt}
\newcommand{\postertablemargin}{$\,$}

\newcommand{\insertRealResultsPoster}{
\begin{table}
{\scriptsize
\begin{center}
\begin{tabular}{l|c|c|c|c}
\toprule
$\;$ & Mean (\ml) & Mean (\bpgml) & Paired t-test & Paired Wilcoxon  \\
\midrule
{Bot-1} &\postertablemargin $0.8941 \pm 0.0103$\postertablemargin 
  &\postertablemargin $0.9066 \pm 0.0104$ \postertablemargin & $0.15022$  &
  \postertablemargin 0.23463 \\
{Bot-2} &\postertablemargin $0.7052 \pm 0.0098$\postertablemargin 
  &\postertablemargin $0.7216 \pm 0.0099$ \postertablemargin & $0.08735$  &
  \postertablemargin 0.06286 \\
\bottomrule
\end{tabular}
\vspace{0.1in}
\end{center}
}
\vspace{0.1in}
\end{table}
}

\newcommand{\insertTableGTDL}{
\begin{algorithm}
\textbf{Given:} $\;$ Data $\{(x_i, y_i, r_i)\}_{i=1}^n$,
  step sizes $\alpha', \alpha''$, 
  return coefficient $\lambda$,
  initial $\xi_0$.
\begin{itemize}[leftmargin=0.3in]
\item[\bulletone] Set $\xi\leftarrow\xi_0$, $w\leftarrow \zero$.

\item[\bulletone] For each epoch $k = 1, 2, \dots$
\begin{itemize}
\item [\bullettwo] Set $\Delta\xi\leftarrow \zero$, $\Delta w \leftarrow \zero$.

\item [\bullettwo] For each episode $i=1,\dots, n$
  \begin{itemize}
  \item [\bulletthree] Set $\;\;\rlambdaTpo\leftarrow r_i$,
      $\;\;\glambdaTpo \leftarrow 0$,  $\;\;\qlambdaTpo \leftarrow \zero$
  \item [\bulletthree] $\rhot \leftarrow \policytheta(\ait|\sit)/q(\ait|\sit)$ for
          $t=1,\dots,\Ti$.
  \item [\bulletthree] For each time step in reverse $t = \Ti,\dots, 1$:
    \begin{itemize} [leftmargin=0.3in]
      \item [(a)] $\glambdat \leftarrow \rhot\Big( (1-\lambda) \valfunhatxi(\sitpo) +
                    \lambda\rhot\rlambdatpo\Big)$
      \item [(b)] $\qlambdat \leftarrow \rhot\Big( (1-\lambda)
        \nabla_\xi\valfunhatxi(\sitpo) + \lambda \qlambdatpo\Big)$
      \item [(c)] $\delta_t \leftarrow \glambdat - \valfunhatxi(\sit)$
      \item [(d)] $h_t \leftarrow \big(\delta_t - w^\top\nabla_\xi\valfunhatxi(\sit)\big)
                  \nabla^2_\xi \valfunhatxi(\sit)\cdot w$
      \item [(e)] $\Delta w \leftarrow \Delta w +  \frac{1}{\Ti}\big(\delta_t -
                  w^\top\nabla_\xi\valfunhatxi(\sit)\big)\nabla_\xi\valfunhatxi(\sit)$
      \item [(f)] $\Delta\xi \leftarrow \Delta\xi + \frac{1}{\Ti}
             \left( \delta_t \nabla_\xi\valfunhatxi(\sit)- 
          \qlambdat w^\top\nabla_\xi\valfunhatxi(\sit) - h_t \right)
                $ 
    \end{itemize} 
  \end{itemize} 
  \item[\bullettwo] $w \leftarrow w + \alpha''\Delta w$.
  \item[\bullettwo] $\xi \leftarrow \xi + \alpha'\Delta\xi$.
\end{itemize} 
\end{itemize}
\caption{\gtdl
\label{alg:gtdl}}
\end{algorithm}
}

\newcolumntype{L}[1]{>{\raggedright\let\newline\\\arraybackslash\hspace{0pt}}m{#1}}
\newcommand{\forceindent}{\leavevmode{\parindent=1em\indent}}

\newenvironment{myindentpar}[1]%
  {\begin{list}{}%
          {\setlength{\leftmargin}{#1}}%
          \item[]%
  }
  {\end{list}}

\definecolor{greycust}{gray}{0.4}
\definecolor{greyagent}{gray}{0.2}
\newcommand{\inqualfont}[1]{{{\fontfamily{cmtt}\selectfont{#1}}}}
\newcommand{\inqualtitlefont}[1]{{{\fontfamily{lmss}\selectfont{\textbf{#1}}}}}
\newcommand{\utterance}[3]{\textcolor{#1}{{\inqualtitlefont{\textbf{#2}}$\;\,$\inqualfont{#3}}}}
\newcommand{\insertscore}[2]{\hfill\textcolor{#1}{\inqualtitlefont{\textbf{[#2]}}}}
\newcommand{\custutt}[1]{\utterance{greycust}{Customer:\hspace{9pt}}{#1}}
\newcommand{\agentutt}[1]{\utterance{greyagent}{Agent:\hspace{23.7pt}}{#1}}
\newcommand{\befbotutt}[2]{\utterance{blue}{\ml:\hspace{34.9pt}}{#2}
\insertscore{blue}{#1}}
\newcommand{\aftbotutt}[2]{\utterance{red}{\ml+\bpg:\hspace{8.0pt}}{#2}
\insertscore{red}{#1}}

\newcommand{\botsep}{0.03in}
\newcommand{\convbotsep}{0.06in}
\newcommand{\bottomspace}{0.04in}
\newcommand{\inserttopspace}{$\;$\\[-0.07in]}
\newcommand{\inserttopspaceshort}{$\;$\\[-0.11in]}

\newcommand{\insertQualResults}{
\begin{table}
\begin{center}
\small
\begin{tabular}{|L{5.35in}|}

\midrule
\inserttopspaceshort
\custutt{Hi there, I'm looking for a restaurant recommendation.} \\
\agentutt{How many in your party?} \\
\custutt{5, including children.} \\[\convbotsep]
\befbotutt{0}{How many people will be joining you?} \\[\botsep]
\aftbotutt{1.5}{That is a fun size, and what is your budget joining you?}\\[\bottomspace]

\midrule
\inserttopspaceshort
\inqualfont{\dots} \\
\custutt{Hey\dots$\,$ Sorry, it's the anniversary of my wife's death. I always do oriental
tonight.} \\
\agentutt{I'm sorry to hear that. I will do my best to find just what you need.} \\
\custutt{Thank you very much.} \\
\agentutt{What kind of budget are you looking at?} \\
\custutt{Doesn't matter... \$100 sounds like a good number.} \\
\agentutt{Will it be just you, or are others going to join you?} \\
\custutt{It's just me tonight, but if you are off soon I wouldn't mind a friendly ear. No
pressure.} \\[\convbotsep]
\befbotutt{0.33}{What type of budget are you looking to stay within?} \\[\botsep]
\aftbotutt{1.33}{How about: Caf\'e Hungry <UNK> ? They serve sea American to oriental on
site such?} \\[\bottomspace]

\midrule
\inserttopspace
\custutt{I would love to find a great Mexican restaurant!} \\
\agentutt{Great, and will anybody be joining you?} \\
\custutt{There will be a total of 3 of us.} \\
\inqualfont{\dots} \\
\custutt{Could you find me a place with outdoor seating?} \\
\agentutt{How about: The Painted Burrito. They serve Mexican food.} \\
\custutt{Eh, I've heard that one isn't great. Can you please suggest another?} \\[\convbotsep]
\befbotutt{0.33}{How about: The Hungry .w seafood restaurant? they serve sea and American
food.} \\[\botsep]
\aftbotutt{1.67}{I am happy to, one moment. How about: Diva Indian Bistro Boston bar? They
serve Mexican food.}\\[\bottomspace]

\midrule
\inserttopspace
\custutt{Hello, I'm looking for a restaurant.} \\
\agentutt{I can help! Please tell me what kind of foods you most enjoy. I see you are in
the Cambridge area.}\\
\custutt{I would like a restaurant that serves Italian food.} \\
\agentutt{Are you looking for something a little less expensive or are you looking for a
high end restaurant?} \\
\custutt{I am within a budget, 20 to 50 dollars.} \\
\agentutt{I highly recommend The Redhouse Restaurant. It serves \dots} \\
\custutt{Is there another restaurant? the restaurant would also need to host a party size
of 5 people.} \\[\convbotsep]
\befbotutt{0}{How about: Amuse? They serve French and American food.} \\[\botsep]
\aftbotutt{2}{How about: Ocean prime - Boston Seaport? They serve sea, American and
Italian food.}\\[\bottomspace]

\midrule
\inserttopspace
\inqualfont{\dots} \\
\agentutt{How about Cafe at Taj Boston?} \\
\custutt{Oh, I just ate there last night. Anything else?} \\
\agentutt{Try Mela Restaurant at 578 Tremont St. Anything else I can help you with today?} \\
\custutt{That sounds nice. Do they allow smoking?} \\[\convbotsep]
\befbotutt{0.5}{This sure is within your budget for all.} \\[\botsep]
\aftbotutt{1.5}{No.}\\[\bottomspace]

\midrule
\end{tabular}
\vspace{-0.15in}
\end{center}
\caption{\small
\label{tb:qualone}
Some qualitative results on the AMT experiment using restaurant data.
In each cell, the text in black/grey show the history of a conversation between the
customer and a human agent.
The dots $\dots$ mean that the part of the conversation history is hidden.
The blue response is by the bot trained using maximum likelihood (\ml) alone.
The red response is by the bot improved with \bpgs after \ml.
The numbers on the right of the blue/red responses are the average scores each response
received from 2-3 AMT workers.
}
\vspace{-0.1in}
\end{table}
}

\newcommand{\insertQualResultsTwo}{
\begin{table}
\begin{center}
\small
\begin{tabular}{|L{5.35in}|}

\midrule
\inserttopspace
\agentutt{How may I help you?} \\
\custutt{I'd like to find a French restaurant.} \\
\agentutt{Do you have a budget in mind?} \\
\custutt{Between \$50 and \$75.} \\
\agentutt{How many will be in your party?} \\
\custutt{Just me.}\\[\convbotsep]
\befbotutt{0}{And how will you be dining?} \\[\botsep]
\aftbotutt{2}{How will you be traveling to the restaurant?}\\[\bottomspace]

\midrule
\inserttopspace
\agentutt{Hello, how may I help you?} \\
\custutt{We are kinda hungry, any suggestion in mind?} \\
\agentutt{What type of cuisine do you like?} \\
\custutt{I want them Americano food.}\\[\convbotsep]
\befbotutt{1}{Very evening. Will you be dining alone?} \\[\botsep]
\aftbotutt{2}{Excellent choice. How many people will be in your party?}\\[\bottomspace]

\midrule
\inserttopspace
\custutt{I'd like to find a good local restaurant.} \\
\agentutt{What type of cuisine would you like?} \\
\custutt{I have a hankering for Indian curry.} \\
\inqualfont{\dots} \\
\agentutt{Will you be needing parking?} \\
\custutt{We'll be driving altogether, so yes.} \\[\convbotsep]
\befbotutt{0}{Thank you for 1. How about: T. w your curry this place should be dining?} \\[\botsep]
\aftbotutt{2}{How about: The Maharaja? They serve Indian and American
food.}\\[\bottomspace]

\midrule
\inserttopspace
\agentutt{How can I help you today?} \\
\custutt{I would like a restaurant recommendation.} \\
\agentutt{Okay what type of resturant?} \\
\custutt{Chinese would be great.} \\
\agentutt{Who will be dining, how will you be travelling?} \\
\custutt{Just me and by bus.} \\[\convbotsep]
\befbotutt{2}{Okay, thanks. Do you have a budget in mind?} \\[\botsep]
\aftbotutt{0}{Do you want a wonderful place evening, or whats your party adults, and
a place.}\\[\bottomspace]

\midrule
\end{tabular}
\vspace{-0.1in}
\end{center}
\caption{\small
\label{tb:qualtwo}
Qualitative results on the AMT experiment using restaurant data, continued.
See caption under Table~\ref{tb:qualone} for more details.
}
\vspace{-0.1in}
\end{table}
}


\newcommand{\pbotsep}{0.03in}
\newcommand{\pconvbotsep}{0.06in}
\newcommand{\pbottomspace}{0.04in}
\newcommand{\pinserttopspace}{$\;$\vphantom{T}\\[-0.27in]}
\newcommand{\pinserttopspaceshort}{$\;$\\[-0.11in]}

\definecolor{pgreycust}{gray}{0.3}
\definecolor{pgreyagent}{gray}{0.1}
\newcommand{\pinqualfont}[1]{{{\fontfamily{cmtt}\selectfont{#1}}}}
\newcommand{\pinqualtitlefont}[1]{{{\fontfamily{lmss}\selectfont{\textbf{#1}}}}}
\newcommand{\putterance}[3]{\textcolor{#1}{{\pinqualtitlefont{\textbf{#2}}$\;\,$\pinqualfont{#3}}}}
\newcommand{\pinsertscore}[2]{\hfill\textcolor{#1}{\pinqualtitlefont{\textbf{[#2]}}}}
\newcommand{\pcustutt}[1]{\utterance{pgreycust}{Customer:\hspace{12pt}}{#1}}
\newcommand{\pagentutt}[1]{\utterance{pgreyagent}{Agent:\hspace{51.0pt}}{#1}}
\newcommand{\pbefbotutt}[2]{\utterance{blue}{\ml:\hspace{79.9pt}}{#2}
\insertscore{blue}{#1}}
\newcommand{\paftbotutt}[2]{\utterance{red}{\ml+\bpg:\hspace{8.0pt}}{#2}
\insertscore{red}{#1}}

\newcommand{\insertQualResultsPoster}{
\begin{table}
\begin{center}
\tiny
\begin{tabular}{|L{12.0in}|}

\midrule
\pinserttopspace
\pcustutt{Hi there, I'm looking for a restaurant recommendation.} \\
\pagentutt{How many in your party?} \\
\pcustutt{5, including children.} \\[\pconvbotsep]
\pbefbotutt{0}{How many people will be joining you?} \\[\pbotsep]
\paftbotutt{1.5}{That$\;$is$\;$a$\;$fun$\;$size,$\;$and$\;$what$\;$is$\;$your$\;$budget$\;$joining you?\hspace{-0.0in}}\\[\pbottomspace]

\midrule
\pinserttopspace
\pcustutt{Hello, I'm looking for a restaurant.} \\
\pagentutt{I can help! Please tell me what kind of foods you most enjoy. I see you are in
the Cambridge area.}\\
\pcustutt{I would like a restaurant that serves Italian food.} \\
\pinqualfont{\dots} \\
\pcustutt{Is there another restaurant? the restaurant \dots} \\[\pconvbotsep]
\pbefbotutt{0}{How about: Amuse? They serve French and American food.} \\[\pbotsep]
\paftbotutt{2}{How about: Ocean prime - Boston Seaport? They serve sea, American and
Italian food.}\\[\pbottomspace]

\midrule
\pinserttopspace
\pinqualfont{\dots} \\
\pagentutt{How about Cafe at Taj Boston?} \\
\pcustutt{Oh, I just ate there last night. Anything else?} \\
\pagentutt{Try Mela Restaurant at 578 Tremont St. Anything else \dots} \\
\pcustutt{That sounds nice. Do they allow smoking?} \\[\pconvbotsep]
\pbefbotutt{0.5}{This sure is within your budget for all.} \\[\pbotsep]
\paftbotutt{1.5}{No.}\\[\pbottomspace]

\midrule
\end{tabular}
\vspace{0.1in}
\end{center}
\vspace{0.1in}
\end{table}
}

\maketitle

\begin{abstract}
\vspace{-0.01in}
We study reinforcement learning of chatbots with recurrent neural network
architectures when the rewards are noisy and expensive to
obtain. For instance, a chatbot used in automated customer service support can
be scored by quality assurance agents, but this process can be expensive, time consuming
and noisy.
Previous reinforcement learning work for natural language processing  uses on-policy
updates and/or is designed for on-line learning settings.
We demonstrate empirically that such strategies are not appropriate for this setting
and develop an off-policy batch policy gradient method (\bpg).
We demonstrate the efficacy of our method via a series of 
synthetic experiments and an Amazon Mechanical Turk experiment on
a restaurant recommendations dataset.
\end{abstract}

\section{Introduction}
\label{sec:intro}

Chatbots are one of the classical applications of artificial intelligence
and are now ubiquitous in technology, business and everyday life.
Many corporate entities are now increasingly using chatbots to either replace  or 
assist humans in customer service contexts.
For example, Microsoft is currently actively building a chat bot to optimise
and streamline its technical support service.

In these scenarios, there is usually an abundance of historical data
since past conversations between customers and human customer service agents
are usually recorded by organisations.
An apparently straightforward solution would be to train chatbots to reproduce
the responses by human agents using standard
techniques such as maximum likelihood.
While this seems natural, it is far from desirable for several reasons.
It has been observed that such procedures have a tendency to produce very generic
responses~\citep{sordoni2015neural}.
For instance, when we trained chatbots via maximum likelihood on a restaurant
recommendations dataset, they  repeatedly output responses to the effect of 
\tokenchars{How large is your group?}, \tokenchars{What is your budget?} etc.
Further, they also produce responses such as 
\tokenchars{Let me look that up.}
or \tokenchars{Give me a second.} which, although permissible for a human agent to
say, are not appropriate for a chatbot.
Although there are ways to increase the diversity of responses~\citep{li2015diversity},
our focus is on encouraging the bot to meaningfully advance the conversation.
One way to address this problem is to provide some form of weak supervision
for responses generated by a chatbot.
For example, a human labeller, such as a quality assurance agent, could score
each response generated by a chatbot in a conversation with a customer.
This brings us to the reinforcement learning (RL) paradigm where these rewards (scores)
are to be used to train a good chatbot. 
In this paper we will use the terms score, label, and reward interchangeably.
Labelled data will mean conversations which have been assigned a reward of some
form  as explained above.

Nonetheless, there are some important differences in the above scenario
when compared to the more popular approaches for RL.
\begin{itemize}[leftmargin=0.3in]
\item \textbf{Noisy and expensive rewards}:
Obtaining labels for each conversation can be time consuming and economically expensive.
As a result, there is a limited amount of labelled data available.
Moreover, labels produced by humans are invariably noisy due to human error and
subjectivity.
\item \textbf{Off-line evaluations:}
Unlike conventional RL settings, such as games, where we try to find the optimal
policy while interacting with the system, the rewards here are not immediately available.
Previous conversations are collected, labelled by human experts, and
then given to an algorithm which has to manage with the data it has.
\item \textbf{Unlabelled Data:}
While labelled data is limited, a large amount of unlabelled data is available.
\end{itemize}
If labelled data is in short supply, reinforcement learning could be
hopeless. However, if unlabelled data can be used to train a decent initial bot, say
via maximum likelihood,
we can use \emph{policy iteration} techniques to refine this bot by making local
improvements using the labelled data~\citep{bellman1956dynamic}.
Besides chatbots, this framework also finds applications in tasks such 
as question answering~\citep{hermann2015teaching,sachan2016science,ferrucci2010watson},
generating image descriptions~\citep{karpathy2015deep}
and machine translation~\citep{bahdanau2014neural}
where a human labeller can provide weak supervision in the form of a score
to a sentence generated by a bot.

To contextualise the work in this paper,
we make two important distinctions in policy iteration methods in reinforcement
learning.
The first is on-policy vs off-policy. In on-policy settings, the goal is to improve
the current policy on the fly while exploring the space.
On-policy methods are used in applications where it is necessary to be competitive
(achieve high rewards) while simultaneously exploring the environment.
In off-policy, the environment is explored using a behaviour policy,
but the goal is to improve a different target policy.
The second distinction is on-line vs batch (off-line).
In on-line settings one can interact with the environment.
In batch methods, which is the setting for this work, one is given past
exploration data from
possibly several behaviour policies and the goal is to improve a target
policy  using this data.
On-line methods can be either on-policy or off-policy whereas batch methods are
necessarily off-policy.

In this paper, we study reinforcement learning in batch settings, for improving
chat bots with Seq2Seq recurrent neural network (RNN) architectures.
One of the challenges when compared to on-line learning is that we do not have interactive
control over the environment.
We can only hope to do as well as our data permits us to.
On the other hand, the batch setting affords us some luxuries. We can reuse existing
data and use standard techniques for hyper-parameter tuning based on cross validation. 
Further, in on-line policy updates, we have to
be able to ``guess'' how an episode will play out, i.e. actions the behaviour/target
policies would take in the future and corresponding rewards.
However, in batch learning, the future actions and rewards are directly
available in the data.
This enables us to make more informed choices when updating our policy.

\subsection*{Related Work}

Recently there has been a surge of interest in deep learning approaches to reinforcement
learning, many of them adopting Q-learning,
e.g.~\citep{mnih2013playing,he2015deep,narasimhan2015language}.
In Q-learning, the goal is to estimate the \emph{optimal action value function} $Q^*$.
Then, when an agent is at a given state, it chooses the best greedy action
according to $Q^*$.
While Q-learning has been successful in several applications,
it is challenging in the settings we consider since estimating $Q^*$
over large action and state spaces will require a vast number of samples.
In this context, policy iteration methods are more promising since we
can start with an
initial policy and make incremental local improvements using the data we have.
This is especially true given that we can use maximum likelihood techniques to estimate
a good initial bot using unlabelled data.

Policy gradient methods, which fall within the paradigm of policy iteration,
make changes to the parameters
of a policy along the gradient of a desired objective~\citep{sutton1999policy}.
Recently, the natural language processing (NLP) literature has turned its attention
to policy gradient methods for improving language models.
\citet{ranzato2015sequence} present a method based on 
the classical REINFORCE algorithm~\citep{williams1992reinforce} for improving
machine translation after preliminary training with maximum likelihood objectives.
\citet{bahdanau2016actor} present an actor-critic method also for machine translation.
In both cases, as the reward, the authors use the BLEU (bilingual evaluation understudy)
score of the output and the translation in the training dataset.
This setting, where the rewards are deterministic and cheaply computable,
does not reflect
difficulties inherent to training chatbots where labels are noisy and expensive.
\citet{li2016deep} develop a policy gradient method bot for chatbots.
However, they use user defined rewards (based on some simple rules)
which, once again, are cheaply obtained and deterministic. 
Perhaps the closest to our work is that of \citet{williams2016end} who use a REINFORCE
based method for chat bots. We discuss the differences of this and other methods
in greater detail in Section~\ref{sec:comparison}.
The crucial difference between all of the above efforts and ours is that they use
on-policy and/or on-line updates in their methods.

The remainder of this manuscript is organised as follows. In Section~\ref{sec:prelims} we
review Seq2Seq models and Markov decision processes (MDP) and describe our framework
for batch reinforcement learning.
Section~\ref{sec:polgrad} presents our method \bpgs and compares
it with prior work in the RL and NLP literature.
Section~\ref{sec:experiments} presents experiments on a synthetic task and a
customer service dataset for restaurant recommendations.

\section{Preliminaries}
\label{sec:prelims}

\subsection{A Review of Seq2Seq Models}
\label{sec:s2s}

The goal of a Seq2Seq model
in natural language processing is to produce an output
sequence $y = [a_1, a_2, \dots, a_T]$  given an input sequence
$x$~\citep{cho2014learning,sutskever2014sequence,kalchbrenner2013recurrent}. 
Here $a_i \in \Acal$ where $\Acal$ is a vocabulary of words.
For example, in machine translation from French to English,
 $x$ is the input sequence in French, and $y$ is its translation in English.
In customer service chatbots, $x$ is the conversation history until the
customer's last query and $y$ is the response by an agent/chatbot.
In a Seq2Seq model, we use an encoder network to represent the input sequence
as a euclidean vector and then a decoder network to convert this vector to an
output sequence.
Typically, both the encoder and decoder networks are recurrent neural networks
(RNN)~\citep{mikolov2010rnn}
where the recurrent unit processes each word in the input/output sequences one at a time.
In this work, we will use the LSTM (long short term memory)~\citep{hochreiter1997lstm}
as our recurrent unit due to its empirical success in several applications.

In its most basic form, the decoder RNN can be interpreted as assigning a probability
distribution over $\Acal$ given the current ``state''. At time $t$, the state $\st$
is the input sequence $x$ and the words $\ytmo = [a_1,\dots,a_{t-1}]$ produced by
the decoder thus far, i.e. $\st = (x, \ytmo)$.
We sample the next word $\at$ from this probability distribution $\policy(\cdot|\st)$,
then update our state $\stpo = (x, \yt)$ where $\yt = [\ytmo,\at]$,
and proceed in a similar fashion.
The vocabulary $\Acal$ contains an end-of-statement token \eostoken.
If we sample \eostokens at time $T+1$, we terminate the sequence and output $\yT$.

\subsection{A Review of Markov Decision Processes (MDP)}
\label{sec:mdp}

We present a formalism for MDPs simplified to our setting.
In an MDP, an agent takes an action $a$ in a state $s$ and transitions to a state
$s'$. An \emph{episode} refers to a sequence of transitions
$s_1\rightarrow a_1 \rightarrow s_2 \rightarrow a_2 \rightarrow \dots \rightarrow
a_{T} \rightarrow \stt{T+1}$
until the agent reaches a terminal state
$\stt{T+1}$. At a terminal state, the agent receives a reward.
Formally, an MDP is the triplet $(\Scal, \Acal, R)$. 
Here, $\Scal$ is a set of states and $\Acal$ is a set of actions.
When we take an action $a$ at state $s$ we transition to a new state $s' = s'(s,a)$
which, in this work, will be deterministic.
$\Acal$ will be a finite but large discrete set and $\Scal$ will be
discrete but potentially infinite.
$R:\Scal\rightarrow\RR$ is the expected reward function such that when we receive a
reward $r$ at state $s\in\Scal$, $\EE[r] = R(s)$.
Let $\termstates\subset\Scal$ be a set of terminal states. When we transition to any
$s\in\termstates$, the episode ends.
In this work, we will assume that the rewards are received only at a terminal state,
i.e $R(s)$ is nonzero only on $\termstates$.

A policy $\policy$ is a rule to select an action at a given state.
We will be focusing on stochastic policies $\policy:\Acal\times\Scal\rightarrow \RR_+$
where $\policy(a|s)$ denotes the probability an agent will execute action $a$ at state
$s$.
We define the \emph{value function} $\valfunpi:\Scal\rightarrow\RR$ of policy $\pi$,
where $\valfun(s)$ is the expected reward at the end of the episode when we follow
policy $\policy$ from state $s$.
For any terminal state $s\in\termstates$, $\valfunpi(s) = R(s)$ regardless of $\pi$.
We will also find it useful to define the \emph{action-value function}
$\actvalfunpi:\Scal\times\Acal:\rightarrow \RR$, where 
$\actvalfunpi(s,a)$ is the expected reward of taking action $a$ at state $s$ and then
following policy $\pi$. With deterministic state transitions this is simply
$\actvalfunpi(s,a) = \valfunpi(s'(s,a))$.
It can be verified that $\valfunpi(s) =
\EE_{a\sim\policy(\cdot|s)}\left[\actvalfunpi(s,a)\right]$~\citep{sutton1998rl}.

\subsection{Set Up}
\label{sec:setup}

We now frame our learning from labels scenario for RNN chatbots as an MDP.
The treatment has similarities to some recent RL work in the NLP literature
discussed above.

Let $x$ be the input and $\ytmo = [a_1,\dots,a_{t-1}]$ be the
words output by the decoder until time $t$.
The state of our MDP at time $t$ of the current episode will be  $\st = (x, \ytmo)$.
Therefore, the set of states $\Scal$ will be all possible pairs of inputs and partial
output sequences. 
The actions $\Acal$ will be the vocabulary.
The terminal states 
$\termstates$ will be $(x, y) $ such that the last literal of $y$ is \eostoken.
The stochastic policy $\pi$ will be a Seq2Seq RNN which produces a distribution over
$\Acal$ given state $\st$.
When we wish to make the dependence of the policy on the RNN parameters $\theta$
explicit, we will write $\policytheta$.
When we sample an action $\at\sim\policy(\cdot|\st)$, we deterministically transition
to state $(x,[\ytmo, \at])$.
If we sample $a_{T+1} = \text{\eostoken}$ at time $T+1$,
the episode terminates and we observe a stochastic reward.

We are given a dataset of input-output-reward triples $\{(\xii{i},\yii{i},\ri)\}_{i=1}^n$
where  $\yi = (\aitt{1},\dots,\aitt{T_i}, \eostoken)$ is the sequence of output words.
This data was collected from possibly multiple \emph{behaviour policies} 
which output $\yii{i}$ for the given input $\xii{i}$.
In the above customer service example, the behaviour policies could be
chatbots, or even humans, which were used for conversations with a customer.
The rewards $r_i$ are scores assigned by a human quality assurance agent
to each response of the chatbot.
Our goal is to use this data to improve  a given target policy $\policytheta$.
We will use $q$ to denote the distribution of the data.
$q(s)$ is the distribution of the states in the dataset,
$q(a|s)$ is the conditional distribution of an action given a state,
and $q(s,a) = q(s)q(a|s)$ is the joint distribution over states and actions.
$q$ will be determined by the initial distribution of the inputs $\xxi$ and
the behaviour policies used to collect the training data.
Our aim is to find a policy that does well with respect to $q$.
Specifically, we wish to maximise the following objective,
\begin{align*}
J(\theta) = \sum_{s\in\Scal} q(s) \valfunpt(s).
\numberthis \label{eqn:pgobj}
\end{align*}
Here, the value function $\valfunpt$ is not available to us but has
to be estimated from the data.
This is similar to objectives used in on-line off-policy policy gradient literature
where $q$ is
replaced by the limiting distribution of the behaviour policy~\citep{degris2012offpac}.
In the derivation of our algorithm, we will need to know $q(a|s)$ to compute the
gradient of our objective. In off-policy reinforcement learning settings this is
given by the behaviour policy which is readily available.
If the behaviour policy if available to us, then we can use this too.
Otherwise, a simple alternative is to ``learn'' a behaviour policy.
For example, in our experiments we used an RNN trained using the unlabelled data
to obtain values for $q(a|s)$. As long as this learned policy can capture
the semantics of natural language (for example, the word \tokenchars{apple} is more
likely than \tokenchars{car} when the current state is $(x, \text{\tokenchars{I ate an}})$),
then it can be expected to do reasonably well.
In the following section, we will derive a stochastic gradient descent (SGD)
procedure that will approximately minimise~\eqref{eqn:pgobj}.

Before we proceed, we note that it is customary in the RL literature to
assume stochastic transitions between states and
use rewards at all time steps instead of the terminal step.
Further, the future rewards are usually
discounted by a discount factor $\gamma < 1$.
While we use the above formalism to simplify the exposition,
the ideas presented here extend naturally to more conventional settings.

\section{Batch Policy Gradient}
\label{sec:polgrad}

Our derivation follows the blueprint in~\citet{degris2012offpac} who derive an
off-policy on-line actor critic algorithm.
Following standard policy gradient methods, we will aim to update the policy 
by taking steps along the gradient of the objective $\nabla J(\theta)$.
\begin{align*}
\nabla J(\theta)
  = 
  \nabla \EE_{s\sim q}\bigg[
    \sum_{a\in\Acal} \policytheta(a|s) \actvalfunpt(s, a)
  \bigg]
  = 
  \EE_{s\sim q}\bigg[
    \sum_{a\in\Acal} \nabla \policytheta(a|s)  \actvalfunpt(s, a)
                      + \policytheta(a|s)  \nabla \actvalfunpt(s, a)
  \bigg].
\end{align*}
The latter term inside the above summation is difficult to work with, so the first step
is to ignore it and work with the approximate gradient
$g(\theta) = \EE_{s\sim q}[\sum_{a\in\Acal}\nabla\policytheta(a|s)\actvalfunpt(s,a)]
\approx \nabla J(\theta)$.
\citet{degris2012offpac} provide theoretical justification for this approximation in
off policy settings by establishing that
$J(\theta) \leq J(\theta + \alpha g(\theta))$ for all small enough $\alpha$.
Expanding on $g(\theta)$, we obtain:
\begingroup
\allowdisplaybreaks
\begin{align*}
g(\theta) &= \EE_{s\sim q}\left[\sum_{a\in\Acal}
   \policytheta(a|s) \frac{\nabla\policytheta(a|s)}{\policytheta(a|s)}\actvalfunpt(s,a)
    \right]
  \;=\;
  \EE_{\substack{s\sim q \\ a\sim q(\cdot|s)}} \Big[
  \rho(s,a) \psi(a,s) \actvalfunpt(s, a)
  \Big] \\
  &= \EE_{(\st,\at)\sim q(\cdot,\cdot)}
  \left[
  \rho(\st,\at) \psi(\at,\st) (\actvalfunpt(\st, \at) - \valfunpt(\st))
   \right].
  \numberthis \label{eqn:gtheta}
\end{align*}
\endgroup
Here $\psi(a,s) = \frac{\nabla\policytheta(a|s)}{\policytheta(a|s)} = 
\nabla \log \policytheta(a|s)$ is the \emph{score function} of the policy
and $\rho(s,a) = \policytheta(a|s)/q(a|s)$ is the \emph{importance sampling} coefficient.
In the last step, we have used the fact that
$\EE[\policy(a|s) \psi(a|s) h(s)] = 0$ for any function $h:\Scal\rightarrow \RR$
of the current state~\citep{szepesvari2010rl}.
The purpose of introducing the value function $\valfunpt$ is to reduce the variance of the
SGD updates -- we want to assess how good/bad action $\at$ is relative to how well
$\policytheta$ will do at state $\st$ in expectation.
If $\at$ is a good action ($\actvalfunpt(\st,\at)$ is large relative to
$\valfunpt(\st)$), the coefficient of the score function is positive
and it will change $\theta$ so as to assign a higher probability to
action $\at$ at state $\st$.

The $\actvalfunpt, \valfunpt$ functions are not available to us so we will replace
them with estimates.
For $\valfunpt(\st)$ we will use an estimate $\valfunhat(\st)$ -- we will discuss choices
for this shortly.
However, the action value function is usually not estimated in 
RL policy gradient settings to avoid the high sample complexity.
A sensible stochastic approximation for $\actvalfunpt(\st, \at)$
is to use the sum of future rewards from the current state~\citep{sutton1998rl}\footnote{
Note $\actvalfunpt(\st, \at) = \valfunpt(\stpo)$ for deterministic transitions.
However, it is important not
to interpret the term in~\eqref{eqn:gtheta} as the difference in the value function
between successive states. Conditioned on the current time step,
$\valfunpt(\st)$ is deterministic, while $\valfunpt(\stpo)$ is stochastic.
In particular, while a crude estimate suffices for the former, the latter is critical
and should reflect the rewards received during the remainder of the episode.
}.
If we receive
reward $r$ at the end of the episode, we can then use $\actvalfunpt(\st, \at)\approx r$
for all time steps $t$ in the episode.
However, since $q(\at|\st)$ is different
from $\policytheta(\att{t}|\stt{t})$ we will need to re-weight future rewards
via importance sampling $r \prod_{i=t}^T\rho(\stt{i},\att{i})$.
This is to account for the fact that 
an action $a$ given $s$ may have been more likely under
the policy $\policytheta(\cdot|s)$ than it was under $q(\cdot|s)$ or vice versa.
Instead of directly using the re-weighted rewards,
we will use the so called $\lambda$--return which is
a convex combination of the re-weighted rewards and the
value function~\citep{sutton1984temporal,sutton1988learning}.
In our setting, they are defined recursively from the end of the episode $t=T+1$ to
$t=1$ as follows. 
For $\lambda \in (0, 1]$,
\begin{align*}
\rplambdaTpo = r, \hspace{0.5in}
\rplambdat = (1-\lambda)\valfunpt(\stpo) +
\lambda \rho(\st,\at) \rplambdatpo\;\;\;\text{for}\;\, t=T,\dots,1.
\numberthis \label{eqn:rlambdat}
\end{align*}
The purpose of introducing $\lambda$ is to reduce the variance of
using the future rewards alone as an estimate for $\actvalfunpt(\st,\at)$. This is
primarily useful when rewards are noisy.
If the rewards are deterministic, $\lambda=1$ which ignores the value function
is the best choice.
In noisy settings, it is recommended to use
$\lambda<1$ (see Sec 3.1 of~\citep{szepesvari2010rl}).
In our algorithm, we will replace $\rplambdat$ with $\rlambdat$ where $\valfunpt$
is replaced with the estimate $\valfunhat$.
Putting it all together, and letting $\alpha$ denote the step size,
we have the following update rule for the parameters $\theta$ of our policy:
\[
\theta \;\leftarrow\;\;
\theta \,+\, \alpha \rho(\st,\at)\psi(\st,\at)(\rlambdat - \valfunhat(\st)).
\]
In Algorithm~\ref{alg:logpolgrad}, we have summarised the procedure where the updates
are performed after an entire pass through the dataset. In practice, we perform
the updates in mini-batches.

\textbf{An Estimator for the Value Function:}
All that is left to do is to specify an estimator $\valfunhat$ for the value function.
We first need to acknowledge that this is a difficult problem:
$\Scal$ is quite large and for typical applications for this work there might not
be enough data since labels are expensive.
That said, the purpose of $\valfunhat$ in~\eqref{eqn:gtheta},~\eqref{eqn:rlambdat}
is to reduce the variance of our SGD updates
and speed up convergence so it is not critical that this be
precise -- even a bad estimator will converge eventually.
Secondly, standard methods for estimating the value function based on minimising the
projected Bellman error require the second derivatives,
which might be intractable for highly nonlinear parametrisations
of $\valfunhat$~\citep{maei2011gradient}.
For these two statistical and computational reasons,
we resort to simple estimators for $\valfunpt$.
We will study  two options.
The first is a simple heuristic used previously in the RL literature,
namely a constant estimator for $\valfunhat$ which is equal to the mean of all
rewards in the dataset~\citep{williams1992reinforce}. 
The second uses the parametrisation $\valfunhat(s) = \sigma(\xi^\top \phi(s))$ where
$\sigma$ is the logistic function and $\phi(s) \in\RR^d$ is a Euclidean representation
of the state.
For $\valfunhat(s)$ of the above form, the Hessian $\nabla^2_\xi \valfunhat(s)$ can be
computed in $\bigO(d)$ time.
To estimate this value function, we use the \gtdls estimator
from~\citet{maei2011gradient}.
As $\phi(s)$ we will be using the hidden state of the LSTM. The rationale for this is as
follows.
In an LSTM trained using maximum likelihood, the hidden state contains useful information
about the objective.
If there is overlap between the maximum likelihood and reinforcement learning objectives,
we can expect the hidden state to also carry useful information about the RL objective.
Therefore, we can use the hidden state to estimate the value function whose
expectation is the RL objective.
We have described our implementation of \gtdls in Appendix~\ref{app:gtdl} and
specified some implementation details in Section~\ref{sec:experiments}.

\insertTableLPG

\subsection*{Comparison with Other RL Approaches in NLP}
\label{sec:comparison}

Policy gradient methods have been studied extensively in \emph{on policy} settings
where the goal is to improve the current policy on the fly~\citep{williams1992reinforce,
amari1998natural}.
To our knowledge, all RL approaches in Seq2Seq models have also adopted on-policy
policy gradient
updates~\citep{bahdanau2016actor,williams2016end,ranzato2015sequence,li2016deep}.
However, 
on policy methods break down in off-policy settings, because any update must account for
the probability of the action under the target policy.
For example, suppose the behaviour policy took action $a$ at state $s$ and received a low
reward. Then we should modify the target policy $\theta$ so as to reduce
$\policytheta(a|s)$. However, if the target policy is already assigning low probability
to $a|s$ then we should not be as aggressive when making the updates.
The re-weighting $\rho(s,a)$ via importance sampling does precisely this.

A second difference is that we study \emph{batch} RL.
Standard on-line methods are
designed for settings where we have to continually improve the target
while exploring using the behaviour policy.
Critical to such methods are the estimation of future rewards
at the current state and the future actions that will be taken by both
the behaviour and target policies.
In order to tackle this, previous research either ignore future rewards
altogether~\citep{williams1992reinforce},
resort to heuristics to distribute a delayed reward to previous
time steps~\citep{bahdanau2016actor,williams2016end}, or
make additional assumptions about
the distribution of the states
\modifone{such as stationarity of the Markov
process}~\citep{maei2011gradient,degris2012offpac}.
However, in batch settings, the $\lambda$-return from a given
time step can be computed directly~\eqref{eqn:rlambdat} since
the future action and rewards are available in the dataset.
Access to this information provides a crucial advantage over techniques
designed for on-line settings.

\section{Experiments}
\label{sec:experiments}

\textbf{Implementation Details:}
We implement our methods using Chainer~\citep{tokui2015chainer},
and group sentences of the same length together in the same batch to make use of GPU
parallelisation.
Since different batches could be of different length, we do not normalise the gradients
by the batch size as we should take larger steps after seeing more data. 
However, we normalise
by the length of the output sequence to allocate equal weight to all sentences.
We truncate all output sequences to length $64$ and use a maximum batch size of $32$.
We found it necessary to use a very small step size ($10^{-5}$), otherwise the
algorithm has a tendency to get stuck at bad parameter values.
While importance re-weighting is necessary in off-policy settings, it can
increase the variance of the updates, especially when $q(\at|\st)$ is very small.
A common technique to alleviate this problem is to clip  the $\rho(\st,\at)$
value~\citep{swaminathan2015counterfactual}.
In addition to single $\rho(\st,\at)$ values, our procedure has a product
of $\rho(\st,\at)$ values when computing the future rewards~\eqref{eqn:rlambdat}.
The effect of large $\rho$ values is a large weight $\rhot(\rlambdat - \valfunhat(\st))$
for the score function in step (ii) of Algorithm~\ref{alg:logpolgrad}.
In our implementation, we clip this weight at $5$ which controls the variance of the
updates and ensures that a single example does not disproportionately affect the gradient.

\textbf{RNN Design:}
In both experiments we use deep LSTMs with two layers for the encoder and decoder
RNNs. 
The output of the bottom layer is fed to the top layer and in the decoder RNN,
the output of the top layer is fed to a softmax layer of size $|\Acal|$.
When we implement \gtdls to estimate $\valfunpt$ 
we use the hidden state of the bottom LSTM as $\phi(s)$.
When performing our policy updates, we only change  the parameters of the top LSTM
and the softmax layer in our decoder RNN. If we were to change the bottom LSTM too, then
the state representation $\phi(s)$ would also change as the policy changes.
This violates the MDP framework.
In other words, we treat the bottom layer as part of the environment in our MDP.
To facilitate a fair comparison, we only
modify the top LSTM and softmax layers in all methods.
We have illustrated this set up in Fig.~\ref{fig:lstmIllus}.
We note that if one is content with using the constant estimator,
then one can change all parameters of the RNN.

\insertFigLSTM

\subsection{Some Synthetic Experiments on the Europarl dataset}
\label{sec:expsyn}

To convey the main intuitions of our method, we compare our methods against other baselines
on a synthetic task on the European parliament proceedings
corpus~\citep{koehn2005europarl}.
We describe the experimental set up briefly, deferring details to
Appendix~\ref{app:syndetails}.
The input sequence to the RNN was each sentence in the dataset.
Given an input, the goal was to reproduce the words in the input without
repeating words in a list of forbidden words.
The RL algorithm does not explicitly know either goal
of the objective but has to infer it from the \emph{stochastic} rewards assigned 
to input output sequences in the dataset.
We used a training set of $500$ input-output-reward triplets for
the RL methods.

We initialised all methods by maximum likelihood training on
$6000$ input output
sequences where the output sequence was the reverse of the input sequence.
The maximum likelihood objective captures part of the RL objective.
This set up reflects naturally occurring practical scenarios for the algorithm where
a large amount unlabelled data can be used to bootstrap a policy
if the maximum likelihood and reinforcement learning objectives are at least partially
aligned. 
We trained the RL algorithms for $200$ epochs on the training set.
At the end of each epoch, we generated outputs from the policy on test set
of $500$ inputs
and scored them according to our criterion. We plot the test set error against
the number of epochs for various methods in Fig.~\ref{fig:synresults}.

Fig.~\ref{fig:ml} compares $3$ methods: \bpgs with and without maximum likelihood
initialisation and 
a version of \bpgs which does not use importance sampling.
Clearly, bootstrapping an RL algorithm with ML can be advantageous especially
if data is abundantly available for ML training.
Further, without importance sampling, the algorithm is not as competitive
for reasons described in Section~\ref{sec:comparison}.
In all $3$ cases, we used a constant estimator for $\valfunhat$ and $\lambda=0.5$.
The dashed line indicates the performance of ML training alone.
\bpgniss is similar to the algorithms of~\citet{williams2016end,ranzato2015sequence}
except that there, their methods implicitly use $\lambda=1$.

Fig.~\ref{fig:constvsgtdl} compares $4$ methods: \bpgs and its on-line version \opgs
 with constant (\const) and \gtdls estimators for $\valfunhat$.
The on-line  versions of the algorithms are a direct implementation of the method
in~\citet{degris2012offpac} which do not use the future rewards as we do.
The first observation is that while \gtdls is slightly better in the early iterations,
it performs roughly the same as using a constant estimator in the long run.
Next, \bpgs performs significantly better than \opg.
\modifone{
We believe this is due to the following two reasons.
First, the online updates assume stationarity of the MDP.
When this does not hold, such as in limited data instances like ours,
the SGD updates can be very noisy.
Secondly, the value function estimate plays a critical role in the online version.
While obtaining a reliable estimate $\valfunhat$ is reasonable in on-line settings where
we can explore indefinitely to collect a large number of samples,
it is difficult when one only has a limited number of labelled samples.
}
Finally, we compare \bpgs with different choices for $\lambda$ in Fig.~\ref{fig:difflambda}.
As noted previously, $\lambda<1$ is useful
with stochastic rewards, but choosing too small a value is detrimental.
The optimal $\lambda$ value may depend on the problem.

\insertFigSynOne

\subsection{Restaurant Recommendations}
\label{sec:exprestaurant}

We use data from an on-line restaurant recommendation service. Customers log into
the service and chat with a human agent asking recommendations for restaurants.
The agents ask a series of questions such as food preferences, group size
etc. before recommending a restaurant.
The goal is to train a chatbot (policy) which can replace or assist the agent.
For reasons explained in Section~\ref{sec:intro}, maximum likelihood training alone will
not be adequate.
By obtaining reward labels for responses produced by various other bots,
we hope to improve on a bot initialised using maximum likelihood.

\textbf{Data Collection:}
We collected data for RL as follows.
We trained five different  RNN chatbots with different LSTM
parameters via maximum likelihood  on a dataset of $6000$ conversations from this dataset.
The bots were trained to reproduce what the human agent said (output $y$) given the
past conversation history (input $x$).
\modifone{
While the dataset is relatively small, we can still expect our bots to do
reasonably well since we work in a restricted domain.
}
Next, we generated responses from these bots on $1216$ separate conversations and had
them scored by workers on Amazon Mechanical Turk (AMT).
For each response by the bots in each conversation,
the workers were shown the history before the particular response and asked to
score (label) each response on a scale of $0-1-2$.
We collected scores from three different workers for each response and used the mean as
the reward.

\textbf{Policies and RL Application:}
Next, we initialised $2$ bots via maximum likelihood and then used \bpgs to
improve them using the labels collected from AMT.
For the $2$ bots we used the following LSTM hidden state size $H$, word embedding size $E$
and \bpgs parameters.
These parameters were chosen arbitrarily and are different from those of the bots 
used in data collection described above.
\begin{itemize}
\item {Bot-1}: $H=512$, $E=256$. $\quad$
\bpg: $\lambda = 0.5$,$\;$ \gtdls estimator for $\valfunhat$.
\item {Bot-2}: $H=400$, $E=400$. $\quad$
\bpg: $\lambda = 0.5$,$\;$ constant estimator for $\valfunhat$.
\end{itemize}

\textbf{Testing:}
We used a separate test set of $500$ conversations which had a total of more than
$3500$ input-output (conversation history - response) pairs.
For each Bot-1 and Bot-2 we generated responses before and after applying \bpg, totalling
$4$ responses per input.
We then had them scored by workers on AMT using the same set up described above.
The same worker labels the before-\bpgs and after-\bpgs responses from the same bot.
This controls spurious noise effects and allows us to conduct a paired test.
We collected $16,808$ before and after label pairs each for Bot-1 and Bot-2 and
compare them using a paired t-test and a Wilcoxon signed rank test.

\textbf{Results:}
The results are shown in Table~\ref{tb:mturk}.
The improvements on Bot-2 are statistically significant at the $10\%$ level on both tests,
while Bot-1 is significant on the Wilcoxon test.
The large p-values for Bot-1 are due to the noisy nature of AMT experiments and we
believe that we can attain significance if we collect more labels
which will reduce the standard error in both tests.
In Appendix~\ref{app:restaurantdetails} we present some examples of conversation
histories and the responses generated by the
bots before and after applying \bpg. 
We qualitatively discuss specific kinds of issues that we were able to overcome via
reinforcement learning.

\insertRealResultsFeb

\section{Conclusion}
\label{sec:conclusion}

We presented a policy gradient method for batch reinforcement learning to train
chatbots. The data to this algorithm are input-output sequences generated using
other chatbots/humans and stochastic rewards for each output in the dataset.
This setting arises in many applications, such as customer service systems,
where there is usually an abundance of
unlabelled data, but labels (rewards) are expensive to obtain and can be noisy.
Our algorithm is able to efficiently use minimal labelled data to improve chatbots
previously trained through maximum likelihood on unlabelled data.
While our method draws its ideas from previous policy gradient work in the 
RL and NLP literature,
there are some important distinctions that contribute to its success in the settings
of interest for this work.
Via importance sampling we ensure that the probability
of an action is properly accounted for in off-policy updates. 
By explicitly working in the batch setting, we are able to use knowledge of
future actions and rewards to converge faster to the optimum.
Further, we use the unlabelled data to initialise our method and also learn a reasonable
behaviour policy.
Our method outperforms baselines on a series of synthetic and real experiments.

The ideas presented in this work extend beyond chatbots. They  can be used in applications
such as question answering, generating image descriptions and machine translation
where an output sentence generated by a policy is scored by a human labeller to
provide a weak supervision signal.

\subsection*{Acknowledgements}
\vspace{-0.05in}
We would like to thank Christoph Dann for the helpful conversations and Michael Armstrong
for helping us with the Amazon Mechanical Turk experiments.

%

{\small
\renewcommand{\bibsection}{\section*{References\vspace{00.1em}} }
\setlength{\bibsep}{3.0pt}
\bibliography{kky,barto}

\begin{thebibliography}{34}
\providecommand{\natexlab}[1]{#1}
\providecommand{\url}[1]{\texttt{#1}}
\expandafter\ifx\csname urlstyle\endcsname\relax
  \providecommand{\doi}[1]{doi: #1}\else
  \providecommand{\doi}{doi: \begingroup \urlstyle{rm}\Url}\fi

\bibitem[Amari(1998)]{amari1998natural}
Shun-Ichi Amari.
\newblock Natural gradient works efficiently in learning.
\newblock \emph{Neural computation}, 10\penalty0 (2):\penalty0 251--276, 1998.

\bibitem[Bahdanau et~al.(2014)Bahdanau, Cho, and Bengio]{bahdanau2014neural}
Dzmitry Bahdanau, Kyunghyun Cho, and Yoshua Bengio.
\newblock Neural machine translation by jointly learning to align and
  translate.
\newblock \emph{arXiv preprint arXiv:1409.0473}, 2014.

\bibitem[Bahdanau et~al.(2016)Bahdanau, Brakel, Xu, Goyal, Lowe, Pineau,
  Courville, and Bengio]{bahdanau2016actor}
Dzmitry Bahdanau, Philemon Brakel, Kelvin Xu, Anirudh Goyal, Ryan Lowe, Joelle
  Pineau, Aaron Courville, and Yoshua Bengio.
\newblock An actor-critic algorithm for sequence prediction.
\newblock \emph{arXiv preprint arXiv:1607.07086}, 2016.

\bibitem[Bellman(1956)]{bellman1956dynamic}
Richard Bellman.
\newblock {Dynamic programming and Lagrange multipliers}.
\newblock \emph{Proceedings of the National Academy of Sciences}, 42\penalty0
  (10):\penalty0 767--769, 1956.

\bibitem[Borkar(1997)]{borkar1997stochastic}
Vivek~S Borkar.
\newblock Stochastic approximation with two time scales.
\newblock \emph{Systems \& Control Letters}, 29\penalty0 (5):\penalty0
  291--294, 1997.

\bibitem[Cho et~al.(2014)Cho, Van~Merri{\"e}nboer, Gulcehre, Bahdanau,
  Bougares, Schwenk, and Bengio]{cho2014learning}
Kyunghyun Cho, Bart Van~Merri{\"e}nboer, Caglar Gulcehre, Dzmitry Bahdanau,
  Fethi Bougares, Holger Schwenk, and Yoshua Bengio.
\newblock Learning phrase representations using rnn encoder-decoder for
  statistical machine translation.
\newblock \emph{arXiv preprint arXiv:1406.1078}, 2014.

\bibitem[Degris et~al.(2012)Degris, White, and Sutton]{degris2012offpac}
Thomas Degris, Martha White, and Richard~S Sutton.
\newblock Off-policy actor-critic.
\newblock \emph{arXiv preprint arXiv:1205.4839}, 2012.

\bibitem[Ferrucci et~al.(2010)Ferrucci, Brown, Chu-Carroll, Fan, Gondek,
  Kalyanpur, Lally, Murdock, Nyberg, Prager, et~al.]{ferrucci2010watson}
David Ferrucci, Eric Brown, Jennifer Chu-Carroll, James Fan, David Gondek,
  Aditya~A Kalyanpur, Adam Lally, J~William Murdock, Eric Nyberg, John Prager,
  et~al.
\newblock Building watson: An overview of the deepqa project.
\newblock \emph{AI magazine}, 31\penalty0 (3):\penalty0 59--79, 2010.

\bibitem[He et~al.(2015)He, Chen, He, Gao, Li, Deng, and Ostendorf]{he2015deep}
Ji~He, Jianshu Chen, Xiaodong He, Jianfeng Gao, Lihong Li, Li~Deng, and Mari
  Ostendorf.
\newblock Deep reinforcement learning with a natural language action space.
\newblock \emph{arXiv preprint arXiv:1511.04636}, 2015.

\bibitem[Hermann et~al.(2015)Hermann, Kocisky, Grefenstette, Espeholt, Kay,
  Suleyman, and Blunsom]{hermann2015teaching}
Karl~Moritz Hermann, Tomas Kocisky, Edward Grefenstette, Lasse Espeholt, Will
  Kay, Mustafa Suleyman, and Phil Blunsom.
\newblock Teaching machines to read and comprehend.
\newblock In \emph{Advances in Neural Information Processing Systems}, pp.\
  1693--1701, 2015.

\bibitem[Hochreiter \& Schmidhuber(1997)Hochreiter and
  Schmidhuber]{hochreiter1997lstm}
Sepp Hochreiter and J{\"u}rgen Schmidhuber.
\newblock Long short-term memory.
\newblock \emph{Neural computation}, 9\penalty0 (8):\penalty0 1735--1780, 1997.

\bibitem[Kalchbrenner \& Blunsom(2013)Kalchbrenner and
  Blunsom]{kalchbrenner2013recurrent}
Nal Kalchbrenner and Phil Blunsom.
\newblock Recurrent continuous translation models.
\newblock In \emph{EMNLP}, volume~3, pp.\  413, 2013.

\bibitem[Kandasamy et~al.(2017)Kandasamy, Bachrach, Tomioka, Tarlow, and
  Carter]{kandasamy2017bpg}
Kirthevasan Kandasamy, Yoram Bachrach, Ryota Tomioka, Daniel Tarlow, and David
  Carter.
\newblock {Batch Policy Gradient Methods for Improving Seq2Seq Conversation
  Models}.
\newblock Unpublished, 2017.

\bibitem[Karpathy \& Fei-Fei(2015)Karpathy and Fei-Fei]{karpathy2015deep}
Andrej Karpathy and Li~Fei-Fei.
\newblock Deep visual-semantic alignments for generating image descriptions.
\newblock In \emph{Proceedings of the IEEE Conference on Computer Vision and
  Pattern Recognition}, pp.\  3128--3137, 2015.

\bibitem[Koehn(2005)]{koehn2005europarl}
Philipp Koehn.
\newblock Europarl: A parallel corpus for statistical machine translation.
\newblock In \emph{MT summit}, volume~5, pp.\  79--86, 2005.

\bibitem[Li et~al.(2015)Li, Galley, Brockett, Gao, and Dolan]{li2015diversity}
Jiwei Li, Michel Galley, Chris Brockett, Jianfeng Gao, and Bill Dolan.
\newblock A diversity-promoting objective function for neural conversation
  models.
\newblock \emph{arXiv preprint arXiv:1510.03055}, 2015.

\bibitem[Li et~al.(2016)Li, Monroe, Ritter, and Jurafsky]{li2016deep}
Jiwei Li, Will Monroe, Alan Ritter, and Dan Jurafsky.
\newblock Deep reinforcement learning for dialogue generation.
\newblock \emph{arXiv preprint arXiv:1606.01541}, 2016.

\bibitem[Maei(2011)]{maei2011gradient}
Hamid~Reza Maei.
\newblock \emph{Gradient temporal-difference learning algorithms}.
\newblock University of Alberta, 2011.

\bibitem[Mikolov et~al.(2010)Mikolov, Karafi{\'a}t, Burget, Cernock{\`y}, and
  Khudanpur]{mikolov2010rnn}
Tomas Mikolov, Martin Karafi{\'a}t, Lukas Burget, Jan Cernock{\`y}, and Sanjeev
  Khudanpur.
\newblock Recurrent neural network based language model.
\newblock In \emph{Interspeech}, volume~2, pp.\ ~3, 2010.

\bibitem[Mnih et~al.(2013)Mnih, Kavukcuoglu, Silver, Graves, Antonoglou,
  Wierstra, and Riedmiller]{mnih2013playing}
Volodymyr Mnih, Koray Kavukcuoglu, David Silver, Alex Graves, Ioannis
  Antonoglou, Daan Wierstra, and Martin Riedmiller.
\newblock Playing atari with deep reinforcement learning.
\newblock \emph{arXiv preprint arXiv:1312.5602}, 2013.

\bibitem[Narasimhan et~al.(2015)Narasimhan, Kulkarni, and
  Barzilay]{narasimhan2015language}
Karthik Narasimhan, Tejas Kulkarni, and Regina Barzilay.
\newblock Language understanding for text-based games using deep reinforcement
  learning.
\newblock \emph{arXiv preprint arXiv:1506.08941}, 2015.

\bibitem[Ranzato et~al.(2015)Ranzato, Chopra, Auli, and
  Zaremba]{ranzato2015sequence}
Marc'Aurelio Ranzato, Sumit Chopra, Michael Auli, and Wojciech Zaremba.
\newblock Sequence level training with recurrent neural networks.
\newblock \emph{arXiv preprint arXiv:1511.06732}, 2015.

\bibitem[Sachan et~al.(2016)Sachan, Dubey, and Xing]{sachan2016science}
Mrinmaya Sachan, Avinava Dubey, and Eric~P Xing.
\newblock Science question answering using instructional materials.
\newblock \emph{arXiv preprint arXiv:1602.04375}, 2016.

\bibitem[Sordoni et~al.(2015)Sordoni, Galley, Auli, Brockett, Ji, Mitchell,
  Nie, Gao, and Dolan]{sordoni2015neural}
Alessandro Sordoni, Michel Galley, Michael Auli, Chris Brockett, Yangfeng Ji,
  Margaret Mitchell, Jian-Yun Nie, Jianfeng Gao, and Bill Dolan.
\newblock A neural network approach to context-sensitive generation of
  conversational responses.
\newblock \emph{arXiv preprint arXiv:1506.06714}, 2015.

\bibitem[Sutskever et~al.(2014)Sutskever, Vinyals, and
  Le]{sutskever2014sequence}
Ilya Sutskever, Oriol Vinyals, and Quoc~V Le.
\newblock Sequence to sequence learning with neural networks.
\newblock In \emph{Advances in neural information processing systems}, pp.\
  3104--3112, 2014.

\bibitem[Sutton(1988)]{sutton1988learning}
Richard~S Sutton.
\newblock Learning to predict by the methods of temporal differences.
\newblock \emph{Machine learning}, 3\penalty0 (1):\penalty0 9--44, 1988.

\bibitem[Sutton \& Barto(1998)Sutton and Barto]{sutton1998rl}
Richard~S Sutton and Andrew~G Barto.
\newblock \emph{Reinforcement learning: An introduction}, volume~1.
\newblock MIT press Cambridge, 1998.

\bibitem[Sutton et~al.(1999)Sutton, McAllester, Singh, Mansour,
  et~al.]{sutton1999policy}
Richard~S Sutton, David~A McAllester, Satinder~P Singh, Yishay Mansour, et~al.
\newblock Policy gradient methods for reinforcement learning with function
  approximation.
\newblock In \emph{NIPS}, volume~99, pp.\  1057--1063, 1999.

\bibitem[Sutton(1984)]{sutton1984temporal}
Richard~Stuart Sutton.
\newblock \emph{Temporal credit assignment in reinforcement learning}.
\newblock University of Massachusetts, Amherst, 1984.

\bibitem[Swaminathan \& Joachims(2015)Swaminathan and
  Joachims]{swaminathan2015counterfactual}
Adith Swaminathan and Thorsten Joachims.
\newblock Counterfactual risk minimization: Learning from logged bandit
  feedback.
\newblock In \emph{Proceedings of the 32nd International Conference on Machine
  Learning}, pp.\  814--823, 2015.

\bibitem[Szepesv{\'a}ri(2010)]{szepesvari2010rl}
Csaba Szepesv{\'a}ri.
\newblock Algorithms for reinforcement learning.
\newblock \emph{Synthesis lectures on artificial intelligence and machine
  learning}, 4\penalty0 (1):\penalty0 1--103, 2010.

\bibitem[Tokui et~al.(2015)Tokui, Oono, Hido, and Clayton]{tokui2015chainer}
Seiya Tokui, Kenta Oono, Shohei Hido, and Justin Clayton.
\newblock Chainer: a next-generation open source framework for deep learning.
\newblock In \emph{Proceedings of Workshop on Machine Learning Systems
  (LearningSys) in The Twenty-ninth Annual Conference on Neural Information
  Processing Systems (NIPS)}, 2015.

\bibitem[Williams \& Zweig(2016)Williams and Zweig]{williams2016end}
Jason~D Williams and Geoffrey Zweig.
\newblock End-to-end lstm-based dialog control optimized with supervised and
  reinforcement learning.
\newblock \emph{arXiv preprint arXiv:1606.01269}, 2016.

\bibitem[Williams(1992)]{williams1992reinforce}
Ronald~J Williams.
\newblock Simple statistical gradient-following algorithms for connectionist
  reinforcement learning.
\newblock \emph{Machine learning}, 8\penalty0 (3-4):\penalty0 229--256, 1992.

\end{thebibliography}


\begin{thebibliography}{3}
\providecommand{\natexlab}[1]{#1}
\providecommand{\url}[1]{\texttt{#1}}
\expandafter\ifx\csname urlstyle\endcsname\relax
  \providecommand{\doi}[1]{doi: #1}\else
  \providecommand{\doi}{doi: \begingroup \urlstyle{rm}\Url}\fi

\bibitem[Degris et~al.(2012)Degris, White, and Sutton]{degris12offpac}
Thomas Degris, Martha White, and Richard~S Sutton.
\newblock Off-policy actor-critic.
\newblock \emph{arXiv preprint arXiv:1205.4839}, 2012.

\bibitem[Maei(2011)]{maei11gtd}
Hamid~Reza Maei.
\newblock \emph{Gradient Temporal-Difference Learning Algorithms}.
\newblock University of Alberta, 2011.

\bibitem[Szepesv{\'a}ri(2010)]{szepesvari10rl}
Csaba Szepesv{\'a}ri.
\newblock Algorithms for reinforcement learning.
\newblock \emph{Synthesis lectures on artificial intelligence and machine
  learning}, 2010.

\end{thebibliography}
}
\bibliographystyle{iclr2017_conference}

\appendix
\newpage
\section*{\textbf{Appendix}}

\section{Implementation of \gtdl}
\label{app:gtdl}

We present the details of the \gtdls algorithm~\citep{maei2011gradient} to estimate
a value function in Algorithm~\ref{alg:gtdl}.
However, while~\citet{maei2011gradient} give an on-line version we present the
batch version here where the future rewards of an episode are known.
We use a parametrisation of the form
$\valfunhat(s) = \valfunhatxi(s) = \sigma(\xi^\top \phi(s))$ where $\xi\in\RR^d$ is
the parameter to be estimated. $\sigma(z) = 1/(1+e^{-z})$ is the logistic function.

The algorithm requires two step sizes
$\alpha',\alpha''$ below for the updates to $\xi$ and the ancillary parameter $w$.
Following the recommendations in~\citet{borkar1997stochastic},
we use $\alpha'' \ll \alpha$. In our implementations, we used
$\alpha' = 10^{-5}$ and $\alpha'' = 10^{-6}$. 
When we run \bpg, we perform steps (a)-(f) of Algorithm~\ref{alg:gtdl} in
step (iii) of Algorithm~\ref{alg:logpolgrad} and the last two update steps of
Algorithm~\ref{alg:gtdl} in the last update step of Algorithm~\ref{alg:logpolgrad}.

The gradient and Hessian of $\valfunhatxi$ have the following forms,
\begin{align*}
\nabla_\xi \valfunhatxi(s) = \valfunhatxi(s)(1 - \valfunhatxi(s))\phi(s),
\hspace{0.3in}
\nabla^2_\xi \valfunhatxi(s)  = 
\valfunhatxi(s)(1 - \valfunhatxi(s))(1 - 2\valfunhatxi(s)) \phi(s) \phi(s)^\top.
\end{align*}
The Hessian product in step (d) of Algorithm~\ref{alg:gtdl}
can be computed in $O(d)$ time via,
\begin{align*}
\nabla^2_\xi \valfunhatxi(s)\cdot w  = 
\left[
\valfunhatxi(s)(1 - \valfunhatxi(s))(1 - 2\valfunhatxi(s)) (\phi(s)^\top w)\right]
\phi(s).
\end{align*}

\insertTableGTDL

\section{Addendum to Experiments}
\label{sec:appExperiments}

\subsection{Details of the Synthetic Experiment Set up}
\label{app:syndetails}

\newcommand{\rforbid}{r_{\text{f}}}
\newcommand{\fforbid}{p_{\text{f}}}
\newcommand{\rreprod}{r_{\text{r}}}

Given an input and output sequence,
we used the average of five Bernoulli rewards $\text{Bern}(r)$, where the parameter $r$ was
$r = 0.75 \times \rreprod + 0.25 \times \rforbid$.
Here $\rreprod$ was the fraction of common words in the input and output sequences
while $\rforbid = 0.01^{\fforbid}$ where $\fforbid$ is the fraction of forbidden words
in the dataset. As the forbidden words, we used the $50$ most common words in
the dataset.
So if an input had $10$ words of which $2$ were forbidden, an output sequence
repeating $7$ of the allowed words and $1$ forbidden word would
receive an \emph{expected} score of $0.75 \times (8/10) + 0.25 \times 0.01^{(1/8)} =
0.7406$.

The training and testing set for reinforcement learning were obtained as follows.
We trained $4$ bots using maximum likelihood on $6000$ input output
sequences as indicated in Section~\ref{sec:expsyn}.
The LSTM hidden state size $H$ and word embedding size $E$ for the $4$ bots 
were, $(H, E) = (256, 128), (128, 64), (64, 32), (32, 16)$.
The vocabulary size was $|\Acal| = 12000$.
We used these bots to generate outputs for $500$ different input sequences each.
This collection of input and output pairs was scored stochastically as described
above to produce a pool of $2000$  input-output-score triplets.
From this pool we use a fixed set of $500$ triplets for testing across all our
experiments.
From the remaining $1500$ data points, we randomly select $500$ for training for
each execution of an algorithm.
For all RL algorithms, we used an LSTM with $16$ layers and $16$ dimensional word
embeddings.

\subsection{Addendum to the AMT Restaurant Recommendations Experiment}
\label{app:restaurantdetails}

\vspace{0.05in}
\subsubsection*{More Details on the Experimental Set up}

We collected the initial batch of training data for RL as follows:
We trained, via maximum likelihood on $6000$ conversations,
five RNN bots whose LSTM hidden size $H$ and word embedding size $E$ were
$(H, E) = (512, 512), (256, 256), (128, 128), (512, 256), (256, 64)$.
The inputs $x$ were all words from the history of the conversation truncated at
length $64$, i.e. the most recent $64$ words in the conversation history.
The outputs were the actual responses of the agent which were truncated to length $64$.
As the vocabulary we use the $|\Acal| = 4000$ most commonly  occurring words in the
dataset and replace the rest with an \tokenchars{<UNK>} token.

Using the bots trained this way we generate responses on 1216 separate conversations.
This data was sent to AMT workers who were asked to label the conversations on the
 following scale.
\begin{itemize}
\item
\textbf{2}: The response is coherent and appropriate given the history 
and advances the conversation forward.
\item
\textbf{1}: The response has some minor flaws but is discernible and appropriate.
\item
\textbf{0}: The response is either completely incoherent or inappropriate and fails to
advance the conversation forward.
\end{itemize}

\vspace{0.10in}
\subsubsection*{Some Qualitative Results}

In Tables~\ref{tb:qualone} and~\ref{tb:qualtwo} we have presented some examples.
The text in black/grey shows the conversation history, the response in blue is by
the bot trained via maximum likelihood (\ml) alone and in red is the bot after
improvement using our \bpgs reinforcement learning algorithm.

The first two examples of Table~\ref{tb:qualone} present examples where the \mls
algorithm repeated generic questions (on budget, group size etc.) even though they had
already been answered previously. After applying \bpg, we are able to correct such issues,
even though there are some grammatical errors.
In the second, third and fourth example, we see that the \ml+\bpgs bot is able to take
context into consideration well when responding.
For example, the customer asks for oriental/Mexican/Italian food. While the
\mls bot doesn't take this into consideration, the \ml+\bpgs bot is able to
provide relevant answers.
However, in the third example, the name of the restaurant suggests that the food might be
Indian and not Mexican.
In the final example of Table~\ref{tb:qualone} the customer asks a direct
question about smoking. The \mls bot provides an irrelevant answer where as the
\ml+\bpgs bot directly responds to the question.

In some examples, the \mls bot had a tendency to produce sentences that were
grammatically correct but nonsensical, sensible but grammatically incorrect, or just
complete gibberish.
We were able to correct such issues via RL.
The first three examples of Table~\ref{tb:qualtwo} present such cases.
Occasionally the opposite happened.
The last example of Table~\ref{tb:qualtwo} is one such instance.

\insertQualResults
\insertQualResultsTwo

\end{document}